\documentclass[a4paper]{article}
\usepackage{paralist}
\usepackage{tabularx}
\usepackage{algorithm2e}
\usepackage{subcaption}
\usepackage{INTERSPEECH2022}

\title{Does It Affect You? Social and Learning Implications of Using Cognitive-Affective State Recognition for Proactive Human-Robot Tutoring}
\name{Matthias Kraus$^1$, Diana Betancourt$^2$, Wolfgang Minker$^2$}
\address{
  $^1$Augsburg University\\
  $^2$Ulm University}
\email{matthias.kraus@uni-a.de, diana.betancourt@uni-ulm.de, wolfgang-minker@uni-ulm.de}

\begin{document}

\maketitle
\begin{abstract}
 Using robots in educational contexts has already shown to be beneficial for a student's learning and social behaviour. For levitating them to the next level of providing more effective and human-like tutoring, the ability to adapt to the user and to express proactivity is fundamental. By acting proactively, intelligent robotic tutors anticipate possible situations where problems for the student may arise and act in advance for preventing negative outcomes. Still, the decisions of when and how to behave proactively are open questions. Therefore, this paper deals with the investigation of how the student's cognitive-affective states can be used by a robotic tutor for triggering proactive tutoring dialogue. In doing so, it is aimed to improve the learning experience. For this reason, a concept learning task scenario was observed where a robotic assistant proactively helped when negative user states were detected. In a learning task, the user's states of frustration and confusion were deemed to have negative effects on the outcome of the task and were used to trigger proactive behaviour. In an empirical user study with 40 undergraduate and doctoral students, we studied whether the initiation of proactive behaviour after the detection of signs of confusion and frustration improves the student’s concentration and trust in the agent. Additionally, we investigated which level of proactive dialogue is useful for promoting the student’s concentration and trust. The results show that high proactive behaviour harms trust, especially when triggered during negative cognitive-affective states but contributes to keeping the student focused on the task when triggered in these states. Based on our study results, we further discuss future steps for improving the proactive assistance of robotic tutoring systems. 
\end{abstract}

\section{Introduction}
There is a long body of research that demonstrates the benefits of robots in educational contexts. For example, it has been shown that human-robot tutoring can lead to increased learning and improved student social behavior \cite{kanda2004interactive,leyzberg2012physical,kennedy2015robot}. An important aspect of successful (robotic) tutors is to keep their students in a positive emotional or affective state, which has positive effects on engagement \cite{efklides2005effects} and the student-tutor relationship \cite{tiberius1993teacher}. Contrary, negative affective states, such as boredom and frustration, have shown to result in a decreased learning performance \cite{craig2004affect}. Therefore, detecting and adequately responding to affective states has become an important aspect in the design of intelligent tutoring systems. D'Mello and Graesser \cite{d2013autotutor} \textsc{AutoTutor} was one of the first works on integrating the affective student state for adapting the tutoring interaction. It showed promising results for even further increasing a student's learning experience. It could detect the negative-affective states of boredom, confusion, and frustration as well as respond to them utilising empathetic and motivational feedback. Similarly, the works by Woolf et al. \cite{woolf2009affect} and Leite et al. \cite{leite2011modelling} studied the impact of empathetic and emotional responses to negative cognitive-affective states. For example, Woolf et al. \cite{woolf2009affect} responded to student frustration by supporting students to understand failure and use it to move the student forward. Additionally, they proposed alternative activities or more challenging projects, when they detected the student to be bored.    

In this work, we take a different direction by presenting a more proactive approach to utilising cognitive-affective student states for enhancing the human-robot tutoring dialogue. According to D'Mello et al. \cite{d2007toward}, an intelligent tutoring agent needs to overcome a reactive policy for recognising and responding to negative affective states and instead proactively anticipate and prevent the onset of such states that harm a student's learning and engagement. They argue that proactively handling a student's cognitive-affective state may optimise learning, by keeping them in positive states. For this reason, we present a novel approach that includes the detection of the onset of the negative cognitive-affective states confusion, and frustration which then triggers the initiation of proactive task-support dialogue for improving the learning experience. Concerning proactivity, we are interested in finding an appropriate level of system intervention. For example, high proactivity may be very useful for explaining learning content, but may also be perceived as obtrusive which could negatively affect user concentration depending on the situation \cite{peng2019design}. Therefore, the level of proactivity needs to be carefully adjusted. For evaluation, we are particularly interested in the impact of our approach on two important aspects of good tutors: supporting the student's learning by keeping the user focused and engaged with the task and expressing trustworthy behaviour for developing an effective human-tutor relationship \cite{saerbeck2010expressive}. In this regard, we investigate the following two research questions:
\begin{enumerate}
    \item Does the initiation of proactive behaviour after the detection of signs of confusion and frustration improve the student's concentration and trust in the agent?
    \item Which level of proactive dialogue is useful for promoting the student's concentration and trust?
\end{enumerate}
For studying these research questions we developed a robotic system based on the \textsc{Nao} robot that was capable of facial expression detection using the \textsc{Affectiva} software \cite{mcduff2013affectiva} for detecting the user's confusion and frustration. In addition, it was able to provide support on four different proactive spoken dialogue levels: It could only react to user requests, provide spoken notifications on how to solve the task, suggest a solution to the student, or automatically solve the task for the student. We conducted an experimental study with the system where study students performed a concept learning task that involved planning, categorising, and decision-making while being accompanied by the robotic tutor. We divided the students into different groups for studying the impact of the different levels of proactive dialogue behaviour. Further, we compared the initiation of proactive dialogue after detecting the signs of confusion and frustration against a control condition as a within-subject experiment. Our main observations from the study results were that high proactive behaviour harms trust, especially when triggered during negative cognitive-affective states but contributes to keeping the user focused on the task when triggered in negative cognitive-affective states. Further, we discuss the limitations of our approach which may be interesting to consider for future research.
\section{Related Work}
\label{sec:relWork}
\subsection{Proactive Human-Robot Interaction}
The concept of proactivity describes self-initiated and anticipatory behaviour to autonomously contribute to problem-solving instead of expressing passive behaviour \cite{nothdurft2015finding}. Due to their highly autonomous nature, robotic applications have become one of the major research streams for studying proactive interaction. Trying to structure the process of a robot's proactive behaviour, Peng et al. \cite{peng2019design} identified three elements: anticipation, initiation of action, and target of impact. The target of impact simply denotes the addressee of the proactive actions. This can be either a human or a group of humans in collaborative task scenarios \cite{wagner2021address,peng2019design}. In this work, we stick to a single user-to-robot interaction. Anticipation refers to the robot's capability to sense its surroundings for predicting future environmental states or the human's intentions. This would subsequently allow a robot to decide when to take anticipatory actions, i.e. to behave proactively. To determine how to physically approach humans acceptably, the user's trajectories and body postures can be used \cite{kato2015may}. However, we do not study physical interaction in this paper and focus on a robot's conversational strategies for providing tutoring.
In this context, a user profile containing the user's location and daytime \cite{grosinger2016making} or the user's motion and speech data \cite{liu2018learning} can be used to make assumptions about the user's behaviour which could then be used for deciding on opportune moments for robot proactivity. Similar to our approach, D'Mello et al. \cite{d2006predicting} estimated the user's cognitive-affective state. Instead of using conversational cues for determining the student's affective state, however, we follow the proposition of Friemel et al. \cite{friemel2018role} who argued that intelligent invocation of user assistance can be provided via measurement of the cognitive-affective states with neuro-physiological tools in real-time. For inferring the states, facial expressions were analysed using the \textsc{Affectiva} software in this work. Providing user-adaptive timing of assistance during tutoring should help to keep the student focused and engaged with the task to achieve increased learning gains as has been shown in related work \cite{leyzberg2014personalizing, rahimi2022timing}.

Initiation of action addresses a robot's autonomy in the functional domains of decision selection and action implementation. Thus, this element is closely related to the robot's level of autonomy during the interaction. Following the categorisation by Sheridan and Verplank \cite{sheridan1978human}, Beer et al. \cite{beer2014toward} described ten levels of robot autonomy (LORA). LORA range from manual teleoperation to full autonomy. At the lower levels, the human-robot interaction (HRI) is generally controlled by the human, however, the robot may assist to some degree with action implementation, e.g., the robot automatically steers to avoid a collision with an obstacle in case a user navigates the robot inappropriately. At the intermediate levels, both interaction partners create plans to achieve a task, however, the human only has supervisory control, whereas the robot takes all actions. At the highest levels, the robot performs all actions of the task with the user only providing a high-level abstract goal. Naturally, higher LORA also allow for a more proactive initiation of interaction, which is essential for cooperative and social robots. Even at a high autonomous level, a robot should interact with the human to some degree due to the human-out-of-the-loop phenomenon in automation that may cause performance problems \cite{endsley1995out}. The proactive initiation of interaction can span various modalities and purposes. For example, a robot can verbally initiate to offer help, if a user runs into problems during executing a task \cite{cramer2009effects}, or making unsolicited suggestions or remarks to the user \cite{peng2019design,liu2018learning, rau2013effects}.

In this work, we utilise a fine-grained model of proactive dialogue for communicating the robot's tutoring behaviour based on a work by Kraus et al. \cite{kraus2020effects} The authors differentiate between four proactive actions which were defined in accordance with their relation to the levels of autonomy: \textit{none}, \textit{notification}, \textit{suggestion}, and \textit{intervention}. These actions range from reactive behavior (``none''-action) to fully proactive system behavior (``intervention''-action) with intermediate levels between those extremes. In several user studies \cite{kraus2020context,kraus2021role}, they found that different levels of proactive dialogue affect the user's perceived trust in the system dependent on specific user traits and the context. Therefore, the level of proactive dialogue is supposed to be important to build a trusting relationship between student and robot, while also keeping the student engaged in the learning task. A detailed description, of how we designed the different levels for tutoring is described in Section 4. 
\subsection{Cognitive-Affective User States}
\begin{figure}
\centering
	\includegraphics[scale = 0.2]{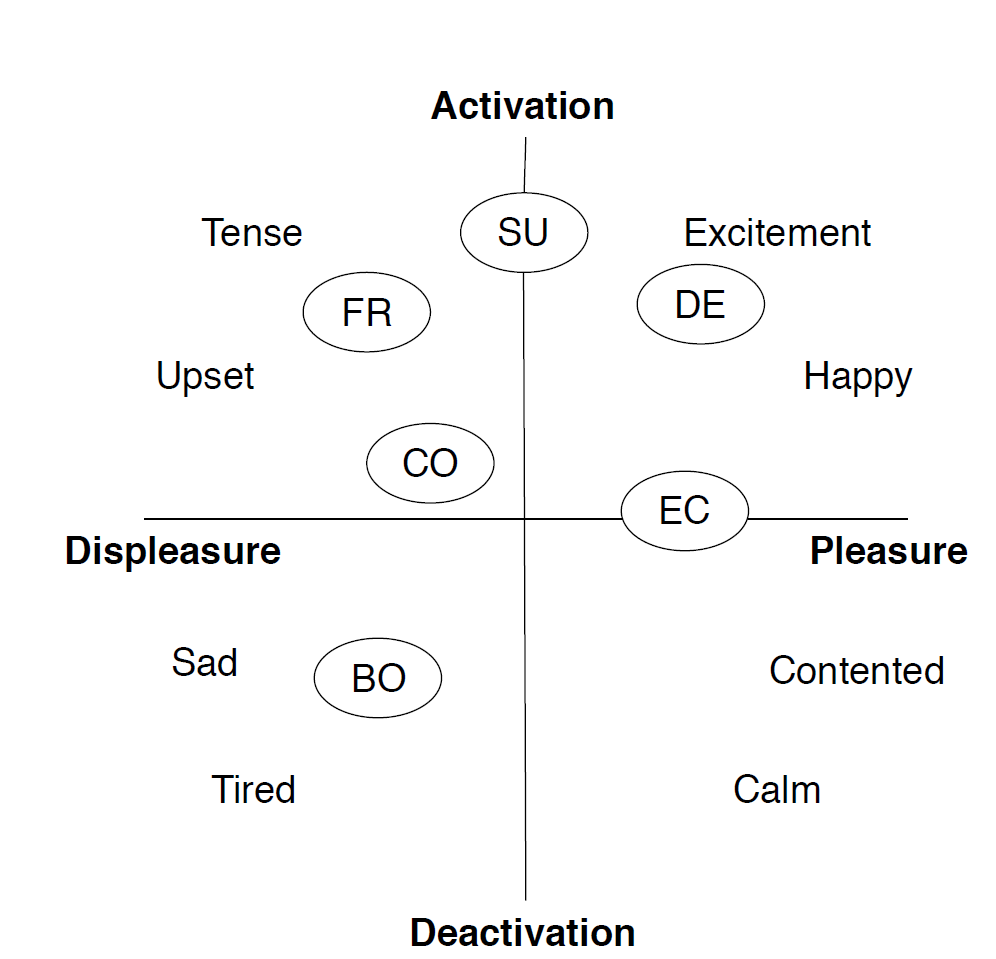}
	\caption{Two-dimensional valence-arousal model showing the position of each cognitive-affective state according to~\cite{baker2010better}: BO - boredom; CO - confusion; DE - delight; EC - engaged concentration/ flow; FR - frustration; SU - surprise.}
	\label{img:cog}
\end{figure}
Cognitive-affective states refer to a human's affective and also cognitive activities \cite{baker2010better}. Certain affects occur predominantly during cognitive activities and learning that are relevant for assistance systems: boredom, frustration, confusion, engagement/flow, delight, and surprise. All cognitive-affective states are situated within the well-known valence-arousal model by Russell \cite{russell2003core} (see Fig.~\ref{img:cog}).  The states are mostly studied via vision, audio and audiovisual recognition \cite{baker2010better}. 
According to Friemel et al. \cite{friemel2018role}, the states that would have the biggest impact on the interaction would be those that depend on the task and also the negative emotions, such as frustration, confusion, or anxiety. Also, Fehrenbacher et al. \cite{fehrenbacher2017information} stated that the mental effort or the cognitive load required to perform the task would influence the need for assistance. Liao et al. \cite{liao2006toward} proposed a dynamic decision framework to unify affect recognition and user assistance. They recognised affective states through active probabilistic inference from multi-modal sensory data. User assistance was then automatically provided through a decision-making process that evaluated the benefits of keeping the user in productive affective states vs. the costs of performing user assistance. Their work focused on affective states that may lead to interaction failures, e.g. stress and fatigue, but not on states that occur during learning, which is the focus of this paper.
Therefore, it is important to define which cognitive-affective states are predominant during learning processes and require tutor intervention. According to the attentional control theory, mainly the user state anxiety would pose a need for assistance \cite{eysenck2007anxiety}. More specific to learning, boredom, frustration, and confusion are relevant for triggering assistance during tutoring. According to D'Mello and Graesser \cite{d2011half}, negative states such as confusion and frustration are usually associated with mistakes, failure, struggling with problems, or revising plans, while positive ones such as excitement or delight are associated with task completion or making discoveries.

We focused on determining the onset of negative states of confusion and frustration for initiating proactive dialogue during human-robot-tutoring. Those negative cognitive-affective states are described by a negative affective valence which can be evaluated by facial expression analysis or with facial electromyography tools \cite{ekman1993facial}. Therefore, indications of these states were estimated using the \textsc{Affectiva} software based on visual cues captured by a webcam. \textsc{Affectiva} has been shown to be a reliable tool for affective state recognition \cite{kulke2020comparison} and provide comparable performance to measurements using EEG \cite{stockli2018facial}.
For evaluating the impact of triggering different levels of proactive tutoring dialogue depending on the student's cognitive-affective state, we focused on two important tutor characteristics: getting the student focused on the learning problem and establishing trust between student and tutor. For measuring the student's focus on the learning task, we measured the user's cognitive load using a standardised questionnaire. Trust was also measured by using student self-reports. The two concepts are explained more in detail in the following.
\subsection{Human-Computer Trust and Cognitive Load}
Trust plays an important role both in interpersonal as well as human-robot relationships \cite{mayer1995integrative, hancock2011meta}. In HRI, trust can be defined as ``the attitude that an agent will help achieve an individual's goal in a situation characterized by uncertainty and vulnerability''~\cite{lee2004trust}. An extensive review of trust in HRI can be found in \cite{hancock2011meta}. 
As there exist various models for human-robot trust, e.g. see Malle and Ullman \cite{malle2021multidimensional}, we refer to the model developed by Madsen and Gregor \cite{madsen2000measuring}. This hierarchical model is built on five basic components of trust: Personal attachment and faith build the bases for affect-based trust while perceived understandability, perceived technical competence, and perceived reliability for cognition-based trust. Affect-based trust refers to long-term human-computer relationships, that can be established through frequent interactions with a system. 

Contrary, cognition-based trust refers to a more short-termed trust. Here, mostly the functionality and usability of a system are of importance. To be perceived as trustworthy, all bases of trust must be perceived as high. Otherwise, the overall trustworthiness could suffer. 

The concept of cognitive load \cite{chandler1991cognitive} connotes the limitation in capacity of the human's working memory when processing the information as well as the limitation in time concerning holding information \cite{paas2010cognitive}. For problem-solving or learning tasks, this implies that material has to be designed in such a way that the usage of the working memory is optimized. Apart from that, the task would congest the cognitive capacity and lead to a sub-optimal outcome. Cognitive load can be distinguished into three independent sources of memory load, namely intrinsic (ICL), extraneous (ECL), and germane cognitive load (GCL) \cite{sweller1998cognitive}. ICL represents the inherent load induced by the content itself. It can not be changed by the learning material and is caused by the complexity/difficulty of the task. ECL arouses from the instructional design of the learning material. This kind of load is linked to mental processes that are not relevant to the task itself, like searching for or narrowing information. GCL is directly linked to the learning process itself. A high germane load indicates that learners are engaged with the task and focus their mental resources on learning processes, e.g. the construction of learning schemes \cite{mayer2002aids}. For measuring whether the user was engaged and focused on the task, we are therefore interested in the measurements of the GCL.
\section{Scenario and System Description}
\begin{figure*}
\centering
   \begin{subfigure}{.49\textwidth}
  \centering
  \includegraphics[width=\linewidth, height=0.33\textheight,keepaspectratio]{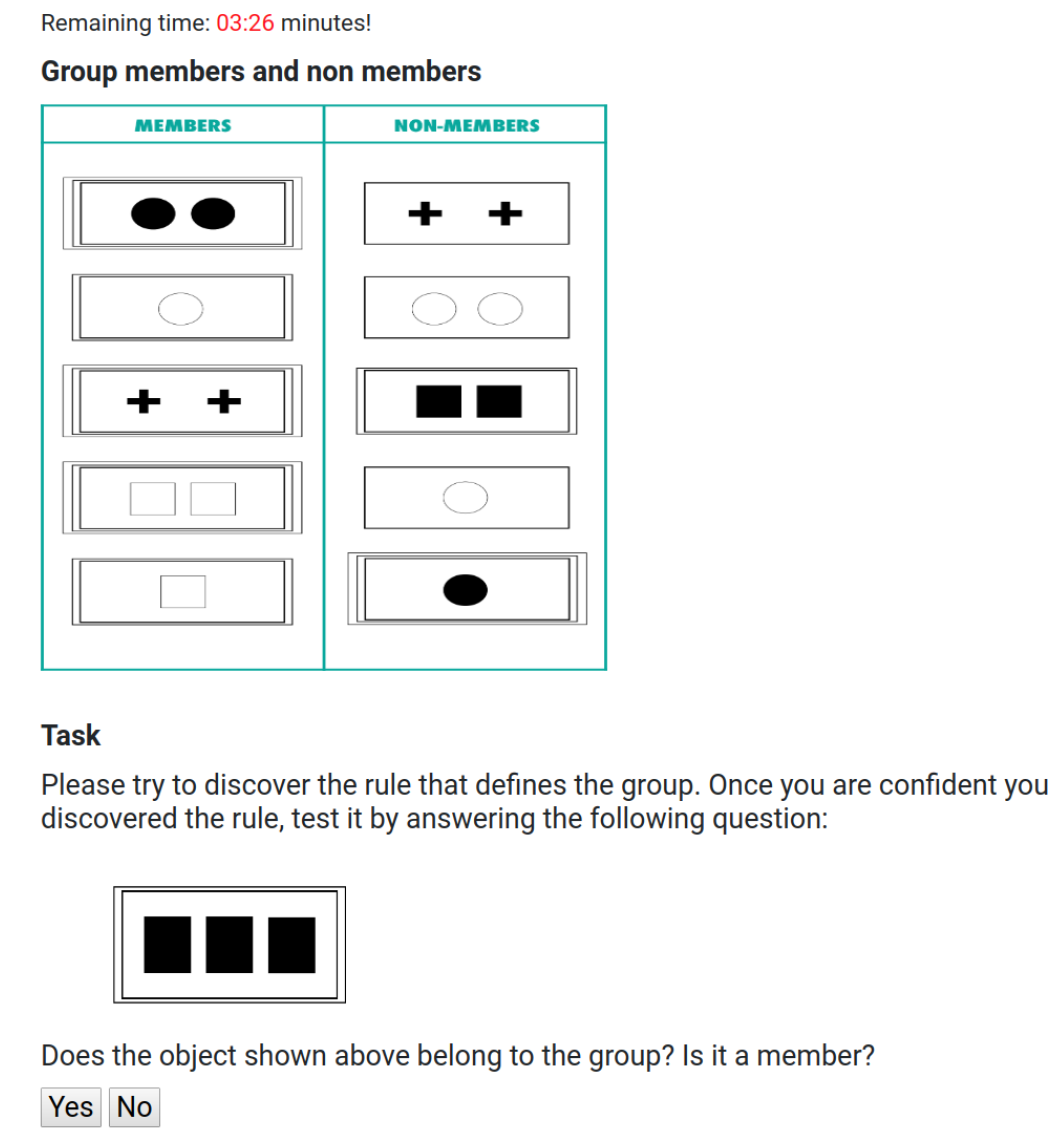}
  \label{fig:sfig1}
    \end{subfigure}
  \begin{subfigure}{.5\textwidth}
  \centering
  \includegraphics[width=\linewidth, height=0.33\textheight]{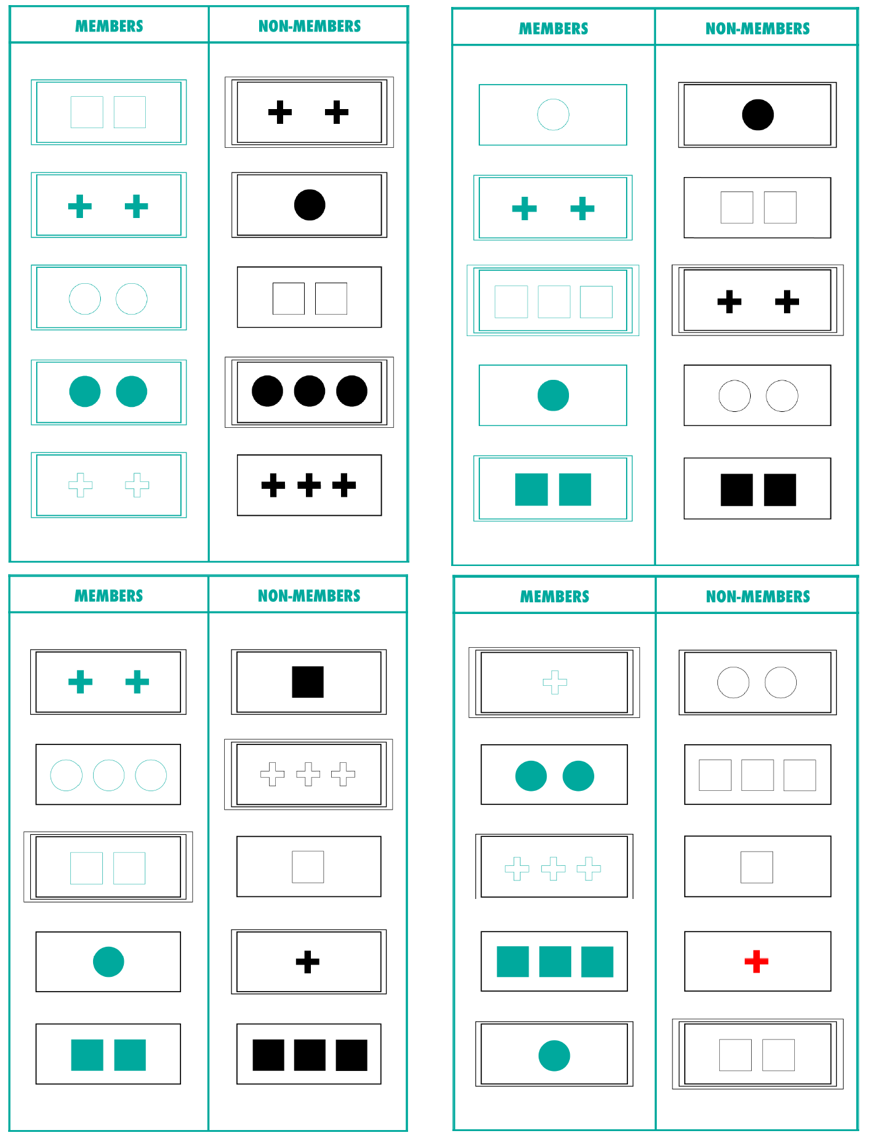}
  \label{fig:sfig2}
    \end{subfigure}
	\caption{Left: Depiction of the user interface for solving the conceptual learning task. Right: Examples of relational rules between objects.}
	\label{img:choice}
\end{figure*} 
For creating a tutoring scenario, a concept learning task was selected in which students were accompanied by a robotic tutoring system. Using a concept learning task was inspired by the work of Bruner et al. \cite{bruner2017study}. Their work was based on how humans categorise information by using a coding system. The students saw ten objects divided into two columns: Five labeled as members, and five as non-members. The task was to deduce the correct rule. Instead of explicitly asking the students for the rule, a new unlabelled object was presented separately. They were asked to classify it as a member or non-member of the group. 

The properties that defined each object were as follows: Number of elements: One, two, or three elements. The shape of elements: A square, a cross, or a circle. The number of borders: One, two, or three borders. Filled or not filled elements. Fig.~\ref{img:choice} (left) illustrates an example of the task presented. This scenario was chosen for two reasons: First, the concept learning task provided a scenario with sufficient complexity where tutoring may be perceived as useful. Secondly, the rule-based structure of the task allowed to equip the assistant with expert knowledge to provide helpful contributions. Each task was limited to four minutes to provide a realistic exam training environment. In the following, the relational rules between objects are described. A visualisation of all possible relations is provided in Fig. \ref{img:choice} (right):
\begin{itemize}
    \item[\textbf{And}] The depiction of the members and non-members in the top left of Fig.\ref{img:choice} (right) exemplifies this rule. In this example, the two properties are two borders and two elements.
    \item[\textbf{Specification}] The two properties are in a certain relation to each other. The depiction of the members and non-members in the top right of Fig.\ref{img:choice} (right) exemplifies this rule. In this example, the number of borders is the same as the number of elements.
    \item[\textbf{Or}]  The depiction of the members and non-members in the bottom left of Fig.\ref{img:choice} (right) exemplifies this rule. All members have two elements in the middle or circles.
    \item[\textbf{Exclusive Or}] The depiction of the members and non-members in the bottom right of Fig.\ref{img:choice} (right) exemplifies this rule. All members either have a cross or a filled element in the middle but not both. The cross in red in the non-member's column refers to an example that has both properties, an element in the middle and a cross, and therefore does not classify as a member.
\end{itemize}
\begin{figure}
\centering
	\includegraphics[width=\columnwidth]{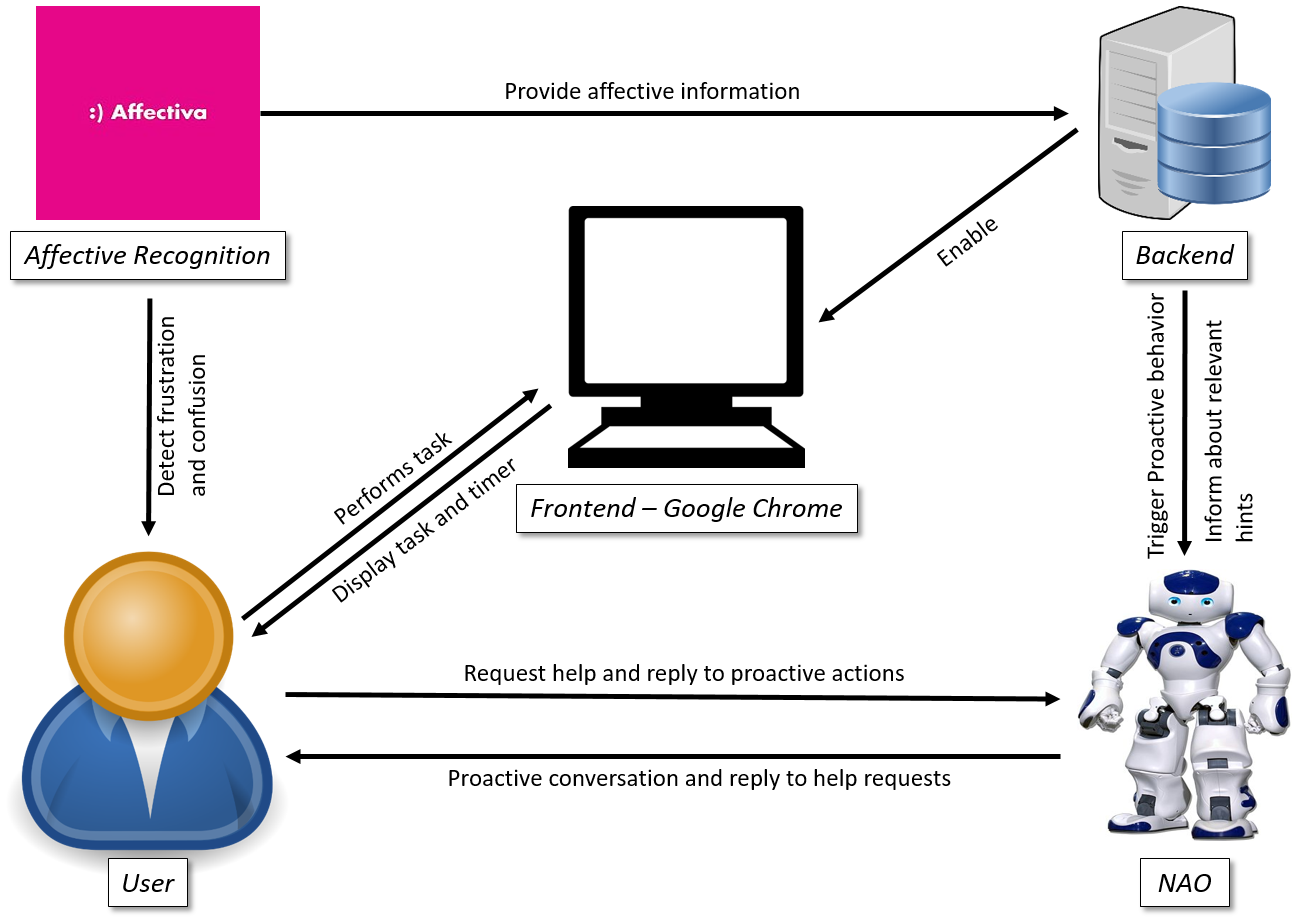}
	\caption{System architecture.}
	\label{img:architecture}
\end{figure}
A Lenovo Thinkpad laptop computer, equipped with a webcam, was used for the experiment. A Tomcat Apache server was used as the back-end for enabling communication between the components. 
The task was presented to students via the Google Chrome web browser. While the students were performing the task on the web browser, assistance was provided through a humanoid robot, \textsc{Nao}, from Aldebaran robotics \footnote{https://www.softbankrobotics.com/emea/en/nao}. The robot was programmed to listen to speech and, if requested, to provide hints for solving the task. This behavior was implemented in the JavaScript QiMessaging API by Aldebaran. \textsc{Nao} received information about task-relevant hints through the back-end. Further, the connection to the back-end allowed to trigger proactive dialogue after a defined amount of time or after the cognitive-affective states of frustration or confusion had been detected. The control timing strategy was implemented with built-in JavaScript functions. For the detection of cognitive-affective states, the \textsc{Affectiva} JavaScript API was used. \textsc{Affectiva} analyses spontaneous facial expressions with facial emotion recognition algorithms. These are trained based on a large database of faces from a variety of different countries and morphological groups. Fig. \ref{img:architecture} shows the interaction between the students and the system on an abstract level.
\section{Design of Proactive Dialogue Strategies}
\subsection{Levels of Proactivity}
As previously mentioned, the proactive tutoring behavior of the robotic assistant was modelled according to leveled proactive dialogue model developed by Kraus et al. \cite{kraus2021role}.
In the following, the individual proactive tutoring strategies are described more in detail:
\subsubsection{None} This level indicated passive tutoring behaviour. Here, the robot reacted to a student`s help request but did not show any proactive behavior. At this level, students could ask for help with any word similar to ``Help.'' or ``Give me hints, please.'', but the system would not autonomously provide hints or solutions. Therefore this level did not require any timing strategy. As shown in the following example, the tutoring would provide a hint after the student's request, which was ought to shift the focus of the student to important aspects of the task:
\begin{compactitem}
    \item[\textbf{Student:}] Nao, help me please.
    \item[\textbf{Nao:}] The rule is of type or.
   \item[\textbf{S:}] Thanks, can you give me another hint?
    \item[\textbf{N:}] Pay attention to the number of elements, it is part of the rule.
     \item[\textbf{S:}] Thank you, NAO.
    \item[\textbf{N:}] You are welcome.
\end{compactitem}
\subsubsection{Notification} This level indicated a more direct way of tutoring. It was implemented by informing that a hint was available, and the students had the option to say whether they wanted to hear it or not. This was ought to provide a more active way of tutoring, but still let the students explore the content themselves by only providing specific details when the user accepted the robot's offer:
\begin{compactitem}
    \item[\textbf{N:}] Help is available. Do you want me to give you a hint?
   \item[\textbf{S:}] Yes, please. / No thanks.
    \item[\textbf{N:}] Focus on the borders. It is relevant to the rule. / Good luck solving the task.
     \item[\textbf{S:}] Thank you, NAO.
    \item[\textbf{N:}] You are welcome.
\end{compactitem}
\subsubsection{Suggestion} This level represented an even more active tutoring approach by automatically suggesting a hint to solve the task and directly providing specific task information. Here, the student could decide whether to use the hint or not, e.g. \textit{``A hint is now available. The rule is of type and''.}
\subsubsection{Intervention} At this level of proactivity, the system took the decisions for the student and directly intervened during the learning process. This behaviour was implemented by the robot providing the task's solution and simultaneously selecting the correct answer on the laptop screen, e.g. \textit{``Help is now available. The rule is: The number of borders is one less than the number of elements. I will choose this option for you''}.

Previous work \cite{rau2013effects,kraus2020effects} found that a low level of proactive dialogue is perceived as more trustworthy than a higher proactive dialogue. Therefore, we hypothesised that the reactive behaviour and notifications are perceived as more trustworthy than directly providing suggestions or even interfering during task execution. As there exists no previous work on the impact of proactive dialogue level on the student's concentration, we provide an exploratory analysis studying this aspect.  
\subsection{Timing Strategies}
The timing strategy determined when the system would take the initiative. For acting upon detected cognitive-affective states an \textsc{Affectiva}-based strategy was used. This strategy would initiate proactive behavior in the detection of the onset of confusion or frustration. The onset of these two states was detected by using facial action units (AU) based on Ekman and Friesen's Facial Action Coding System (FACS) \cite{ekman1993facial}. The FACS allows linking active AUs to the underlying basic emotions of sadness, happiness, surprise, disgust, anger, and fear. Based on this, McDaniel et al. \cite{mcdaniel2007facial} and Craig et al. \cite{craig2008emote} showed that AUs 4 (brow lowered) and 7 (lid tightened) indicate confusion and AUs 1 (inner brow raised) and 2 (outer brow raised) are a manifestation of frustration. The proactive dialogue was triggered when either indication of frustration or confusion was detected by the \textsc{Affectiva} API. The individual steps handled internally by the API were as follows: First, the webcam would provide raw pixel images. Afterward, regions of interest, i.e. pixels of facial information, were extracted using landmark detection. Finally, machine learning regressors predicted activity scores between 0 (no activity) and 1 (high activity) for each AU and returned the results. The predictors were trained on the \textsc{Affectiva} dataset. The activity detected by \textsc{Affectiva} API in each of the AUs was compared to a threshold that would trigger the robot. That threshold was defined in a pre-test with three additional students. In this way, thresholds that seemed to provide a sensible trade-off between precision and recall were determined empirically. 

A strategy using well-defined timing was considered as the control condition. In this condition, three moments were set in each task and a random function was implemented in JavaScript to choose one of these three moments at each task. The moments could be thirty seconds, two minutes and a half, and three minutes. Each task lasted 4 minutes so the intervention would occur in this time-lapse at different moments in each task. 

Concerning related work \cite{d2007toward, friemel2018role}, we argue that triggering proactive depending on the onset of confusion or frustration could help to increase the user's focus on the task and increase the robot's perceived trustworthiness as compared to the control condition.
\section{Experimental Setup}
A factorial 2 x 4 mixed design was used. The independent variables manipulated were: timing strategies triggering proactive dialogue (control condition vs. triggered by the cognitive-affective states confusion/frustration) as within-subject and the levels of proactive dialogue (reactive - notification - suggestion - intervention) as the between-subject factor. Students were randomly distributed to each of the between-subject factors and were confronted with both timing strategies during the study. To minimise sequence effects, the order of the timing strategies was randomised.

The students were welcomed to the study and the initial instructions for the study were presented. The informed consent was given along with any clarification on the adherence to data privacy standards. Students were told they would participate in a study based on HRI in which decision-making skills and cooperation with the robot would be evaluated. Students were also informed they would be recorded via video for further analysis of the interaction. Information about the assistance via proactive dialogue strategies using their facial expression analysis for cognitive-affective states and timing was initially omitted to avoid expectancy effects. However, they received details about the speech capabilities of \textsc{Nao}. After the introduction, a pre-questionnaire was presented including the demographics and possible confounding variables, e.g. negative attitudes towards robots. Afterward, the experiment started. The experiment consisted of two rounds each of which a student had to solve five tasks (10 tasks in total). Per task, they would see 10 objects like those exemplified in Fig.\ref{img:choice}. Additionally, each task had an upper limit of 4 minutes indicated by a timer on the screen. During the experiment, the students were confronted with one of the four proactive dialogue levels. For each round of tasks, the robot's proactive behaviour was triggered using one of the two timing strategies. After the completion of the first round, the timing strategy switched. Further, students had to fill in a questionnaire to assess the dependent variables and to check the manipulation after each round. The used questionnaires are listed in Table \ref{table:questionnaires}. The whole procedure lasted between 45 minutes to one hour. 
\begin{table}
\centering
\begin{tabularx}{\columnwidth}{p{6cm}|X}
\textit{Questionnaires} & \textit{Variable Type} \\
\hline
Trust \cite{jian2000foundations} & Dependent \\ 
Cognition-Based Trust (Competence, Reliability, Understandabiltity) \cite{madsen2000measuring} & Dependent \\
Affect-Based Trust (Faith, Attachment) \cite{madsen2000measuring} & Dependent \\ 
Acceptance \cite{van1997simple}& Dependent \\
Cognitive Load \cite{laugwitz2006konstruktion}& Dependent \\
Usability \cite{brooke1996sus} & Dependent \\
User Experience \cite{laugwitz2006konstruktion} & Dependent \\
Technical Affinity \cite{karrer2009technikaffinitat} & Confounding \\
Negative Attitudes Towards Robots \cite{bartneck2004design}& Confounding \\
Trust Propensity \cite{merritt2013trust}& Confounding
\end{tabularx}
\caption{Questionnaires used. All questionnaires were adapted to a 7-point Likert scale from ``Completely disagree'' to ``Completely agree''.}
\label{table:questionnaires}
\end{table}
40 subjects were recruited for the study. However, three subjects had to be excluded due to not complying with the study guidelines. The average age was 26 years (Std = 5.14). 37 \% were females, while 63 \% percent were males. A high to advanced level of English knowledge was required to perform the study, which was why 72 \% percent of the subjects had a C2 and C1 (expert) level of English. The rest had advanced knowledge. Subjects' English proficiency was self-reported. 60 \% of the subjects were undergraduate students and the remaining were doctoral students. All subjects received a 5€-Amazon-Voucher as compensation independent of the study outcome. 
\section{Results}
\begin{table}
\centering
\begin{tabularx}{\columnwidth}{X||X|X}
\textit{Proactive Action} & \textit{Perceived Activeness} & \textit{T-test compared to None-Action} \\
\hline
None & 0.36 (1.38) & *** \\
Notify & 2.04 (3.39) & p = .296 \\
Suggestion& 1.85 (2.80) & p = .296 \\
Intervention& 3.17 (2.19)  & p = .030 \\
\end{tabularx}
\caption{Manipulation check of the perceived proactivity of the system. Perceived Activeness was measured as the mean of the difference for the rating scales if students perceived the assistant as active and on the scale whether they perceived it as reactive.}
\label{table:manipul}
\end{table}
For data analysis, we used t-tests for the manipulation checks, a multivariate ANOVA for confounding variables, as well as a mixed ANOVA for testing the interaction between the different proactive actions and timing strategies. A Bonferroni-Holm correction was applied, where multiple testing was conducted. No significant outliers were found in the data set. Confounding group differences for proactive behavior could be ruled out as the multivariate ANOVA did not reveal any significant differences (all p-values $>> .05$). The manipulation for proactive dialogue behaviour was successful, as the proactive actions were consistently rated higher than reactive behavior for the user-perceived activeness of the system. However, only the difference between the intervention action and reactive behavior was significant. The means and standard deviations along with the p-values are presented in Table \ref{table:manipul}. 
\subsection{User Experience with the Experimental Setup}
In general, the system received positive feedback. Students accepted their interaction partner ($M = 5.24,~SD = 1.11$) and had a good experience, represented by a high UEQ-value ($M = 5.40,~SD = 0.93$). In addition, the interaction with \textsc{Nao} received moderate ratings for usability ($M = 3.78,~SD = .39$). Generally, students had high trust in the system ($M = 5.63,~SD = .78$) as well as their sub-components reliability ($M = 5.21,~SD = 1.02$), competence ($M = 5.24,~SD = 1.01$), understandability ($M = 5.67,~SD = .93$), and faith ($M = 5.25,~SD = 1.09$). Moderate ratings for personal attachment ($M = 3.98,~SD = 1.37$) were reported.
\subsection{Effects of Proactive Dialogue Strategies on Usability and the Learner's Focus}
Regarding usability, we only found a statistically significant effect of the timing strategies for the \textit{Intervention} strategy. Here, the \textsc{Affectiva}-trigger was rated significantly lower than the control-trigger for usability ($F(1, 8) = 12.34,~p = .027,~\eta^{2} = .61$). For investigating the learner's concentration on the learning, we investigated the GCL, which is correlated with a learner's task engagement and task focus \cite{mayer2002aids}. Considering the \textit{Intervention} strategy, we found the GCL to be rated higher for the \textsc{Affectiva}-trigger ($F(1, 8) = 4.67,~p = .063,~\eta^{2} = .37$).
\subsection{Effects of Proactive Dialogue Strategies on the Learner's Trust}
There was a statistically significant interaction between proactive dialogue and the timing strategies for perceived understandability ($F(3, 34) = 3.45,~p = .027,~\eta^{2} = .23$) and a tendency towards an interaction for personal attachment ($F(3, 34) = 2.51,~p = .076,~\eta^{2} = .18$). For investigating the simple main effects of proactive dialogue and timing strategies, we conducted a one-way, respective repeated measures ANOVA. There were no simple main effects of the proactive dialogue levels depending on the timing strategies (all p-values $>> .05$). However, we found a statistically significant effect of timing for the \textit{Intervention} strategy. The \textsc{Affectiva}-trigger was rated significantly lower than the control-trigger for perceived understandability ($F(1, 8) = 6.40,~p = .035,~\eta^{2} = .44$). Furthermore, we found a tendency towards faith in the system ($F(1, 8) = 3.64,~p = .093,~\eta^{2} = .31$) being increased by the \textit{Intervention} strategy. A significant effect timing on faith in the system was found ($F(1, 34) = 4.46,~p = .042,~\eta^{2} = .12$). Generally, students had more faith in the robot acting according to the control timing condition. Additionally, we found a tendency that the \textsc{Affectiva}-trigger resulted in less perceived system competency ($F(1, 34) = 3.25,~p = .080,~\eta^{2} = .09$)

\begin{figure}
\centering
	\includegraphics[scale=0.3]{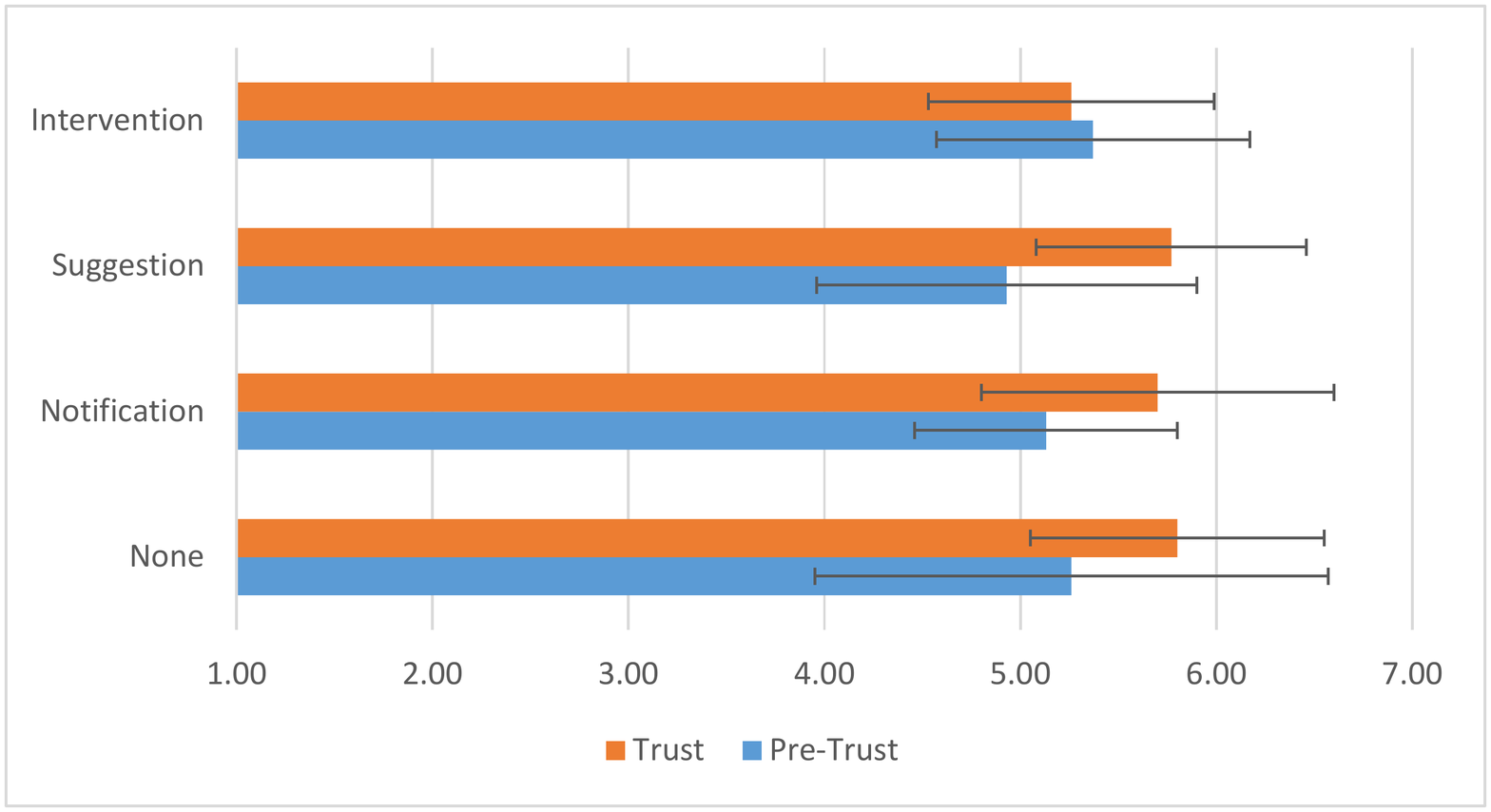}
	\caption{Evolution of the trust developed throughout the experiment concerning the proactive dialogue levels.}
	\label{img:trustevoAffectiva}
\end{figure}

For considering the trust progression throughout the experiment depending on the proactive dialogue level, we investigated the within-subject differences in the trust ratings before and after the experiment. Initial trust was measured using the trust propensity scale. For testing the significance of the differences, we used paired t-tests. Here, we found a significant positive trust development for the \textit{Suggestion} strategy ($t(9) = -4.28,~p = .002$). In summary, reactive and proactive behavior had a positive effect on establishing trust, except for the \textit{Intervention} strategy. The results are depicted in Fig.~\ref{img:trustevoAffectiva}. No significant effects were found for student characteristics, e.g. gender or age.
\begin{table*}
\centering
\begin{tabularx}{\textwidth}{XX||X|X|X|X|X}

      \hline
     \emph{Proactive Action}& &\textbf{Trust} & \textbf{Acceptance}  & \textbf{Usability}& \textbf{UEQ} &  \textbf{GCL} \\
       &  && & & &  \\
      
      & & \textit{M (SD)} & \textit{M (SD)} &\textit{M (SD)} &\textit{M (SD)}  &\textit{M (SD)} \\
      \hline
     \textbf{None} & overall  &5.80 (.75)&  5.46 (.70)& 3.89 (.34)& 5.48 (.96) & 5.19 (.74) \\
                   & Affectiva&5.79 (.74)&  5.43 (.73)& 3.96 (.36)& 5.43 (1.08)& 5.33 (.75) \\
                      & Control &5.81 (.83)&  5.49 (.75)& 3.81 (.43)& 5.54 (.87)& 5.05 (.73) \\
     \hline
     \textbf{Notification}& overall & 5.70 (.90) & 5.30 (1.33) & 3.76 (.45) & 5.54 (1.08)& 4.59 (1.40) \\
                    & Affectiva&5.65 (1.03)&  5.33 (1.53)& 3.71 (.88)& 5.52 (1.06)& 4.61 (1.55)\\
                      & Control &5.75 (.87)&  5.27 (1.16)& 3.81 (.37)& 5.56 (1.11)& 4.56 (1.25)\\
     \hline
      \textbf{Suggestion}& overall&5.77 (.69)& 5.30 (1.06) & 3.90 (.39) & 5.37 (.93)& 4.99 (.96) \\
                    & Affectiva &5.79 (.67)&  5.26 (.96)& 3.89 (.47)& 5.40 (.67)& 5.00 (1.05) \\
                      & Control &5.75 (.85)&  5.33 (1.16)& 3.91 (.31)& 3.34 (1.00)& 4.97 (.84)\\
      \hline
        \textbf{Intervention}& overall& 5.26 (.73) & 4.99 (.79)&  3.63 (.39) & 5.26 (1.04)& 4.97 (.77)\\
                     & Affectiva&5.11 (1.00)&  4.97 (.98)& 3.37 (.47)& 5.06 (1.19) & 5.37 (.72)\\
                      & Control& 5.41 (.64)&  5.02 (0.88)& 3.90 (.44)& 5.47 (.99) & 4.56 (.82)\\
      \hline
\end{tabularx}
\caption{Descriptive statistics of the overall perceived trust and acceptance towards the system, as well as ratings for usability, user experience, and germane cognitive load regarding the proactive dialogue strategies.}
\label{table:descriptiveAffectiva}
\end{table*}
\section{Discussion}
\subsection{High proactive behaviour harms trust especially when triggered during negative cognitive-affective states...}
As shown in Table \ref{table:descriptiveAffectiva}, low- to medium-levels of proactive dialogue (\textit{None, Notification, Suggestion}) received higher ratings for perceived trust than a high-level of proactivity (\textit{Intervention}). This is in line with our hypothesis based on related work showing that a high degree of proactivity has a decreasing effect on trust \cite{rau2013effects,kraus2021role}. 
Further, we considered the trust evolution throughout the experiment depending on the proactive dialogue levels (see Fig. \ref{img:trustevoAffectiva}). According to Glikson and Woolley \cite{glikson2020human}, initial trust starts at a low level and builds over time during interaction with robotic AI. In this regard, we found that the \textit{Reactive}, and \textit{Notification} strategies showed a tendency but only the \textit{Suggestion} strategy significantly increased the perceived user trust. Thus, these trust trajectories seem to be in line with related work. However, for the \textit{Intervention} strategy a trust decrease was found.
Both observations validate the negative effect of high proactivity on the student-tutor trust relationship. This effect is even increased when triggering proactivity depending on the onset of negative cognitive-affective states which may be due to this timing behaviour being less understandable than the control timing condition. Generally, the \textsc{Affectiva}-trigger also showed signs of decreasing the perceived competency of the tutoring robot. Thus, the deteriorated cognition-based trust may be the main reason for the trust decrease of the \textit{Intervention} strategy. This becomes especially evident when observing the affect-based trust represented by the perceived faith in the robot's actions. Faith is only increased using the \textit{Intervention} strategy when triggered at the onset of confusion or frustration, but for all other proactive dialogue strategies faith was higher for the control timing condition. Thus, the loss of trust using the \textsc{Affectiva}-trigger did not stem from the affect-based trust. In summary, the timing of proactive behaviour during tutoring has an important impact on the cognition-based trust relationship and becomes more relevant the more autonomously a system acts.
\subsection{... But contributes to concentration when triggered at negative cognitive-affective states}
Surprisingly, the \textit{Intervention} strategy was the only one to significantly increase a student's rated GCL when triggered during the onset of states of confusion or frustration (see Table \ref{table:descriptiveAffectiva}). A high GCL is related to a learner's engagement with the task and their focus on the learning processes \cite{mayer2002aids}. Even though this strategy had a low impact on the cognitive-based of trust, it positively contributed to a user's concentration. We assume that the low trustworthiness of this proactive dialogue level could explain the student's increased focus. As students did trust the automatic decision of the system less, they may have double-checked the answers of the system more closely and put more thought into the task, which could have resulted in a learning gain. Thus, a competitive robot that presents and selects the correct solution for providing samples to the students when they are frustrated or confused could be beneficial for task concentration. However, our hypothesis that proactive dialogue behaviour based on a negative cognitive-affective state increases the student's concentration could not be proved in general and was only found to be valid for high proactivity.
\subsection{Focusing on negative cognitive-affective states and facial features may be insufficient}
Even though we found some significant differences in observing different kinds of proactive dialogue either triggered after the onset of confusion/frustration or in a controlled manner, only focusing on negative cognitive-affective states was a clear limitation of the presented approach. According to Baker et al. \cite{baker2010better} and Graesser et al. \cite{graesser2007exploring}, people cycle through all kinds of cognitive-affective states during learning. As the relations between the different states are still rather unclear, frustration and confusion may be a natural aspect of the experience of learning when dealing with difficult material and should not be avoided in general. Thus, tutor proactivity is not necessarily always required in such states and solely relying on cognitive-affective states for triggering proactive behaviour may be insufficient. As a consequence, also other student-related information, e.g. level of knowledge, may be utilised in combination with a student's mental state for deciding on proactive behaviour. Furthermore, some challenges that Li and Ji \cite{li2005active} mentioned could be contemplated when considering cognitive-affective states. The first of the challenges are related to the sensory observations often being imprecise and uncertain. Although the Affectiva software showed good reliability for measuring affective states using facial cues \cite{kulke2020comparison, stockli2018facial} there still exist bias and inaccuracies in the data on which the recognition module was trained. To reduce this imprecision, different modalities are suggested to be considered. This was not done in the current study to create a more natural interaction and avoid being intrusive. Subsequently, for future studies, multi-modal affect detection in non-intrusive ways could be considered. Finally, we only used one proactive dialogue level per experiment. As there seems to be a trade-off between increasing the student's perceived trust in robotic assistant (low levels of proactive dialogue) and increasing the student's concentration (high proactivity), the proactive dialogue levels need to be adapted to specific situations and students for being more effectively integrated into useful and trusted robotic tutors.
\section{Conclusion}
We observed the impact of a robot tutor's proactive dialogue triggered at the onset of the negative cognitive-affective states of confusion and frustration on the learner's concentration and the student-tutor trust relationship. In an empirical study using a \textsc{Nao} as a robotic tutor for students working on a concept learning task, we found that high proactive behaviour harms the student's perceived trust in the system. Particularly, when triggered during negative cognitive-affective states. We found that the highest level of proactive dialogue positively contributes to a learner's concentration when triggered during the onset of these states. For this, it may be beneficial to generally look more into the relations between trust in a tutoring system and the user's task concentration which could result in a learning gain. We deem the sole focus on negative-affective states and facial cues for recognition insufficient for invoking proactive dialogue assistance.

We propose to use multi-modal signals for detecting cognitive-affective states and to focus more on the individuality of the students. Modeling the idiosyncrasy of each student would lead to personalised systems by using models that can adapt to the student's characteristics and situation. One example of this is the use of predictive models \cite{bull2010open}. Depending on specific user characteristics, user-adapted proactive behaviour could be more beneficial for creating a trustworthy learning experience.

\bibliographystyle{IEEEtran}

\bibliography{mybib}

\end{document}